\documentclass{article}

\usepackage[final]{corl_2019} 

\usepackage{verbatim}
\usepackage{xfrac}
\usepackage{subfig}
\usepackage{mathtools}
\usepackage{times}
\usepackage{epsfig}
\usepackage{graphicx}
\usepackage{amsmath}
\usepackage{amssymb}
\usepackage{dblfloatfix}
\usepackage{booktabs, multicol, multirow}
\usepackage{combelow}
\usepackage{flushend}
\usepackage{color}    
\definecolor{lightgray}{rgb}{0.95, 0.95, 0.95}
\definecolor{fullred}{rgb}{0.85,.0,.1}

\author{Rare\cb{s} Ambru\cb{s} \quad Vitor Guizilini \quad Jie Li \quad  \quad Sudeep Pillai \quad Adrien Gaidon\\
Toyota Research Institute (TRI) \\
{\tt\ firstname.lastname@tri.global}}

\title{Two Stream Networks for \\ Self-Supervised Ego-Motion Estimation}

\begin{document}
\vspace*{-4mm}
\maketitle
\vspace*{-4mm}


\begin{abstract}
Learning depth and camera ego-motion from raw unlabeled RGB video streams is seeing exciting progress through self-supervision from strong geometric cues.
To leverage not only appearance but also scene geometry, we propose a novel self-supervised two-stream network using RGB and inferred depth information for accurate visual odometry. In addition, we introduce a sparsity-inducing data augmentation policy for ego-motion learning that effectively regularizes the pose network to enable stronger generalization performance. As a result, we show that our proposed two-stream pose network achieves state-of-the-art results among learning-based methods on the KITTI odometry benchmark, and is especially suited for self-supervision at scale. 
Our experiments on a large-scale urban driving dataset of 1 million frames indicate that the performance of our proposed architecture does indeed scale progressively with more data.




\end{abstract}







\linespread{0.88}
\section{Introduction}
\label{sec:intro}

Visual ego-motion estimation is a fundamental capability in mobile robots, used in many tasks such as perception, navigation, and planning. While visual ego-motion estimation has been well-studied in the Structure-from-Motion~\cite{hartley2003multiple} and Visual-SLAM~\cite{davison2007monoslam,murartal2015ORBSLAM} literature, recent work has shown exciting progress in self-supervised learning methods~\cite{zhou2017unsupervised,li2017undeepvo,clark2017vinet,pillai2017towards}. These methods are versatile and scalable, as they learn directly from raw data, typically using the proxy photometric loss as a supervisory signal~\cite{zhou2017unsupervised, godard2017unsupervised}. Similar to~\citet{zhou2017unsupervised}, we jointly learn a monocular depth and camera ego-motion network in a self-supervised manner. While recent works in self-supervised monocular depth and pose estimation have mostly focused on engineering the loss function~\cite{yin2018geonet,zou2018dfnet,mahjourian2018unsupervised,casser2018depth}, we show that performance in this self-supervised SfM regime critically depends on the model architecture, in line with the observations of~\citet{kolesnikov2019revisiting} and \citet{guizilini2019packnet}. 

In this work, we specifically address the current limitations of self-supervised ego-motion learning architectures, namely their exclusive reliance on dense appearance changes, ignoring sparse structures that make the strength of more traditional SfM algorithms. We also investigate the strong interdependence between depth and pose in this self-supervised learning regime. We make three main contributions in this work.
First, we propose \textbf{a novel two-stream network combining images and inferred depth for accurate camera ego-motion estimation}. Our architecture, inspired by action recognition models~\cite{simonyan2014two}, efficiently leverages appearance and scene geometry, reaching state-of-the-art performance among learning-based methods on the KITTI odometry benchmark.

Second, while most learning-based methods tend to rely on generic model regularization policies to avoid overfitting, we introduce \textbf{a sparsity-inducing image augmentation scheme specifically targeted at regularizing camera ego-motion learning}. Through experiments, we show that our aggressive augmentation policy indeed reduces overfitting in this self-supervised regime,
providing a simple-yet-effective mechanism to learn a sufficiently-sparse network for pose estimation. 

Third, we quantify \textbf{the performance benefits and scalability of self-supervised pre-training on large datasets}. We introduce an urban driving dataset of 1 million frames, and show that by pre-training the network with large amounts of data we are able to improve monocular ego-motion estimation performance on a target dataset such as KITTI~\cite{geiger2013vision}. 

\section{Related Work}
\label{sec:related}


\textbf{Self-supervised methods for depth and ego-motion estimation} 
have become popular, as accurate ground-truth measurements rely heavily on more expensive and specialized equipment such as LiDAR and Inertial Navigation Systems (INS).
One of the earliest works in self-supervised depth estimation~\cite{godard2017unsupervised} used the photometric loss as a proxy for supervision to learn a monocular depth network from stereo imagery. In this work, the authors leverage differentiable view-synthesis~\cite{jaderberg2015spatial} to geometrically synthesize the left stereo image from the right image pair and the predicted left disparity, permitting a proxy loss to be imposed between the geometrically synthesized image and the actual image captured in a stereo camera.
~\citet{zhou2017unsupervised} extend this self-supervision to the generalized multi-view case, and leverage constraints typically incorporated in Structure-from-Motion
to simultaneously learn depth and camera ego-motion from monocular image sequences.
Several works have extended this work further - engineering the loss function to handle errors in the photometric loss via flow~\cite{yin2018geonet, teed2018deepv2d, zou2018dfnet}, robustly handling outliers in the loss~\cite{godard2018digging,zhou2018unsupervised}, incorporating 3D constraints~\cite{mahjourian2018unsupervised}, explicitly modelling dynamic object motion~\cite{luo2018every}, and employing stereo and monocular constraints in the same framework~\cite{li2017undeepvo, pillai2018superdepth, yang2018deep}.
~\citet{teed2018deepv2d} proposed an iterative method to regress dense correspondences from pairs of depth frames and compute the 6-DOF estimate using the Perspective-n-Point (PnP)~\cite{lepetit2009epnp} algorithm. Instead, in this work we show that performance in the self-supervised SfM regime critically depends on the choice of the model architecture and the specific ego-motion optimization task at hand. By drawing insights through ablation studies, we introduce a sparsity-inducing image augmentation scheme to effectively regularize ego-motion learning, instead of only limiting such modifications to the underlying loss-function.


\textbf{Multi-Stream architectures for multi-modal learning} 
While recent works in self-supervised SfM learning have focused on tailoring the loss function~\cite{yin2018geonet,zou2018dfnet,godard2018digging}, a few methods following~\cite{kolesnikov2019revisiting} have explored the space of network architectures for such tasks~\cite{pillai2018superdepth,guizilini2019packnet,godard2018digging}.~\citet{godard2018digging} used a novel architecture relying on a ResNet~\cite{he2016deep} backbone which is shared between the depth and the ego-motion network. In the context of multi-modal and multi-task learning, multi-stream architectures have been shown to perform remarkably well in different challenging problems such as object detection and classification~\cite{chen2017multi,chung2017two, zanuttigh2017deep}, semantic segmentation~\cite{hazirbas2016fusenet,valada2018self}, action recognition~\cite{simonyan2014two, carreira2017quo}, and image enhancement~\cite{zhang2018density}. ~\citet{chen2017multi} uses a region proposal branch to construct a zenithal view of the LiDAR point cloud that is then applied over a depth and RGB stream. In~\cite{zanuttigh2017deep}, the authors classify objects by using six separate convolutional branches, each receiving a different depth map view of the object under consideration.~\citet{chung2017two} approach multi-modal learning by feeding two siamese networks with RGB and optical flow features for the task of person re-identification. In other works~\cite{simonyan2014two,zhang2018density,radwan2018vlocnet++}, multi-stream architectures have been designed for multi-task learning where one of the streams guide the learning for the other parallel stream by providing aggregated context and additional conditioning information.~\citet{simonyan2014two} employ a two-stream architecture separately utilizing single frames and multi-frame optical flow over each branch for action recognition and video classification. Following it, in~\cite{carreira2017quo}, authors study different architectures for the same objective, including inflated 3D convolutional ones. Inspired by both multi-task and multi-modal network architectures, we treat the RGB image and predicted monocular depth as two separate input modalities and introduce a two-stream architecture tailored for self-supervised ego-motion learning. Through experiments, we show that the proposed network architecture is able to extract and appropriately fuse RGB-D information from each of its branches to enable accurate ego-motion estimation. 



	
\section{Self-supervised Two-Stream Ego-motion Estimation}
\label{sec:procedure}

\subsection{Method overview}

As originally proposed in their work,~\citet{zhou2017unsupervised} define the task of simultaneously learning depth and pose from a monocular image stream and utilize the proxy photometric loss introduced in~\cite{godard2017unsupervised} to self-supervise both tasks. 
%
%
While~\citet{zhou2017unsupervised} and others~\cite{godard2018digging,zou2018dfnet,mahjourian2018unsupervised,casser2018depth} have mostly limited to operating purely in the RGB domain, we note that depth prediction tasks naturally permit multi-modal and multi-task reasoning for further downstream processing.

\begin{figure}[t]
    \centering
    {
    \includegraphics[ width=1.\linewidth, height=0.38\linewidth]{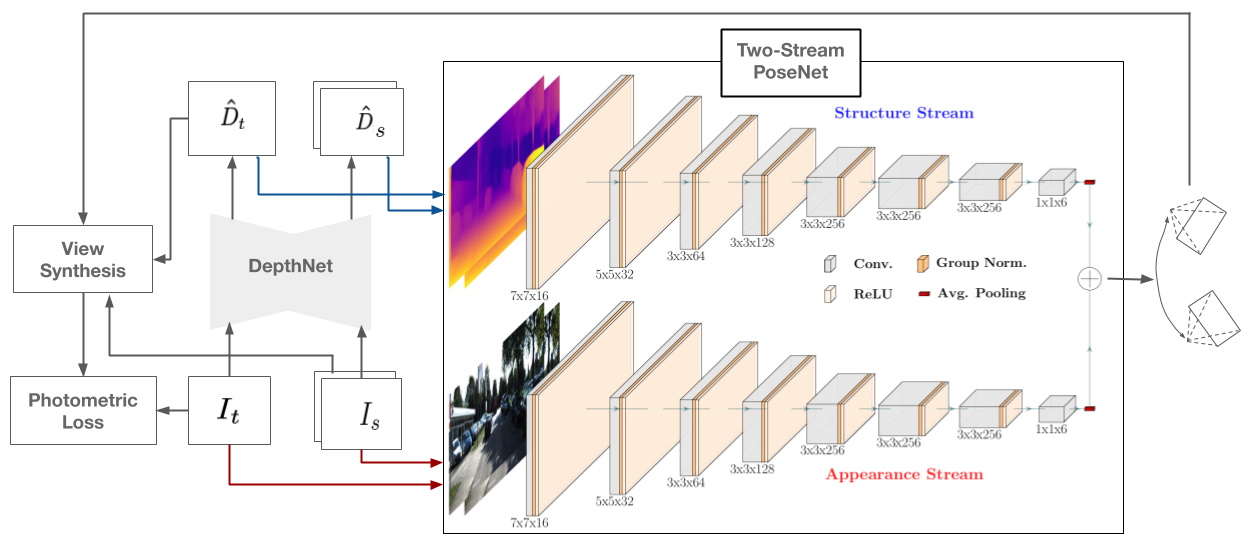}
    }
    \caption{Our proposed self-supervised depth and camera ego-motion learning method. Contrary to previous self-supervised depth and pose estimation methods~\cite{zhou2017unsupervised,zou2018dfnet}, the RGB ($I_s, I_t$) and predicted depth ($\hat{D}_s, \hat{D}_t)$ images from the source and target frames are used in two separate network streams in our modified PoseNet architecture. The resulting two-stream pose network is self-supervised, achieving state-of-the-art performance on the KITTI odometry benchmark~\cite{geiger2013vision}.}
    \label{fig:sfm-architecture}
\end{figure}

To this end, we extend the formulation in~\cite{zhou2017unsupervised} to consider the fusion of RGB and depth information within the pose network, via a two-stream network architecture.
%
Figure~\ref{fig:sfm-architecture} illustrates our overall self-supervised learning method with architecture details of our two-stream pose network.
Our modified pose network $f_{\mathbf{x}}: (I_t,D_t,I_S,D_S) \to \mathbf{x}_{t \to S}$ estimates the 6-DOF ego-motion transformation $\mathbf{x}_{t \to s} = \begin{psmallmatrix}\mathbf{R} & \mathbf{t}\\ \mathbf{0} & \mathbf{1}\end{psmallmatrix} \in \text{SE(3)}$ between target frame $I_t$ and (temporally adjacent) source frames $I_s$ for $s\in S$, with the additional predicted depth information as inputs. 
This allows the proposed pose network to effectively fuse multi-modal RGB-D information for the task of ego-motion estimation, and is able to further decompose the task beyond reasoning over raw input image streams, as typically done in~\cite{zhou2017unsupervised,zou2018dfnet,li2017undeepvo}.
Our depth estimation network $f_d: I \to D$ is based on the DispNet network architecture~\cite{mayer2016large}, which is a baseline commonly used in the literature~\cite{zhou2017unsupervised,pillai2018superdepth}. The network employs a decoder with skip connections from the encoder's activation blocks and outputs depths at 4 scales. Depth at each scale is upsampled by a factor of 2 and concatenated with the decoder features to help resolve the depth and the next scale. A more detailed description  of the our depth estimation network can be found in the appendix. 



\subsection{Two-stream ego-motion network}
\label{subsec:pose_network}

The proposed architecture for the ego-motion estimation task is shown in Figure~\ref{fig:sfm-architecture}. Inspired by work in action recognition~\cite{carreira2017quo,simonyan2014two} we propose to augment the commonly used ego-motion estimation network which relies only on RGB input  ~\cite{zhou2017unsupervised,godard2017unsupervised,mahjourian2018unsupervised} with a second modality by passing the estimated depth, along with the RGB, as inputs to the network. This modification allows the network to learn both appearance and geometry features, leading to better results. 

The architecture consists of two towers, each processing one of the modalities as shown in Figure~\ref{fig:sfm-architecture}. Each tower contains 8 convolutional layers plus a final average pooling layer and outputs the 6-DOF transformation between the input frames. The 6-DOF transformation is represented as 6 numbers $\left(x,y,z\right)$ for the translation and $\left(\alpha, \beta, \gamma \right)$ for the rotation using the Euler parameterization. We experimented with other parameterizations for the rotation (e.g. log-quaternion~\cite{brahmbhatt2018geometry}, or quaternion~\cite{kendall2017geometric}) but did not see any improvement.

We note that unlike related methods in the literature which process multiple frames (most methods usually process 3 frames to estimate ego-motion), our network is designed to process only two frames at a time. We show through experiments that this simple architecture is powerful enough to capture the complex dynamics of outdoor environments. 

\subsection{Self-supervised objective for depth and ego-motion learning}

Following~\cite{zhou2017unsupervised}, we formulate the self-supervised objective as the minimization of the proxy photometric loss imposed between the target image $I_t$ and the synthesized target image $\hat{I}_t$ generated from the source view. Notably, both the depth and pose networks are trained jointly via the same self-supervised proxy measure, rendering the two tasks strongly coupled with each other. 

The overall loss is composed of a robust appearance loss term estimated via Structural Similarity~\cite{wang2004image}, and a depth regularization term~\cite{godard2017unsupervised}. The robustness incorporated helps account for errors incurred due to occlusions and dynamic objects~\cite{godard2018digging}.
 
\textbf{Robust appearance-based loss} Following~\cite{godard2017unsupervised,zhou2017unsupervised,pillai2018superdepth,guizilini2019packnet} we define the appearance-based matching loss between two images as the linear combination between an $L1$ loss and the Structural Similarity (SSIM) loss~\cite{wang2004image} given by:
\begin{equation}
    \mathcal{L}_{p}\left(I_t,\hat{I_t}\right) = \alpha~\frac{1 - \text{SSIM}\left(I_t,\hat{I_t}\right)}{2} + \left(1-\alpha\right)~\| I_t - \hat{I_t} \|
  \label{eq:loss-photo}
\end{equation}
The SSIM component of the loss is further described in the appendix. The photometric loss as defined in Equation~\ref{eq:loss-photo} is susceptible to errors induced by occlusions or dynamic objects. While the authors in~\cite{zhu2018robustness} suggest clipping the photometric errors above a percentile to filter out errors, in practice we found that the \textit{auto-masking} approach of~\cite{godard2018digging} yields better results. Given target image $I_t$, source image $I_s$, and synthesized image $\hat{I_t}$ we define the masking term $\mathcal{M}_{r}$:
\begin{equation}
    \mathcal{M}_{r}\left(I_t,I_s,\hat{I_t}\right) = \mathcal{L}_{p}\left(I_t,I_s\right) < \mathcal{L}_{p}\left(I_t,\hat{I_t}\right) 
  \label{eq:automask}
\end{equation}
$\mathcal{M}_{r}$ is a robust mask that allows us to filter out stationary pixels and pixels with little photometric variation~\cite{godard2018digging}. Finally, we define the robust appearance matching loss between target image $I_t$ and context images $I_S$ as:
\begin{equation}
    \mathcal{L}_{r}\left(I_t,I_S\right) = \min_{s\in S} \mathcal{M}_{r}\left(I_t,I_s,\hat{I_t}\right)\cdot\mathcal{L}_{p}\left(I_t,\hat{I_t}\right)
  \label{eq:loss-robust}
\end{equation}
We note that $\mathcal{L}_{r}$ is a per-pixel loss - we denote the term $\mathcal{M}_{r}$ that filters out static pixels between $I_t$ and $I_s$, and subsequently select the loss term with the lowest value across all context images in $I_S$.

\textbf{Depth smoothness loss} In addition to the photometric term, we incorporate a multi-scale edge-aware term to regularize the depth in texture-less regions~\cite{godard2017unsupervised}:
\begin{equation}
  \mathcal{L}_{s}(\hat{D}_t) = | \delta_x \hat{D}_t | e^{-|\delta_x I_t|} + | \delta_y \hat{D}_t | e^{-|\delta_y I_t|}
  \label{eq:loss-disp-smoothness}
\end{equation}
Finally, the loss we optimize is:
\begin{equation}
    \mathcal{L}(I_t,I_S) = \mathcal{L}_r(I_t,I_S) \odot \mathcal{M}_r +  \lambda~\mathcal{L}_s(\hat{D}_t)
    \label{eq:overall-loss}
\end{equation}
Prior to computing $\mathcal{L}$, we upsample the depth maps across all the scales to the resolution of the input image $I_t$, following insights from~\cite{pillai2018superdepth,zhu2018robustness,godard2018digging}.

\subsection{Sparsity-inducing data augmentation for ego-motion estimation}
\label{subsec:sparsity-augmentation}

\begin{figure}[t]
\centering
    \includegraphics[width=0.16\columnwidth]{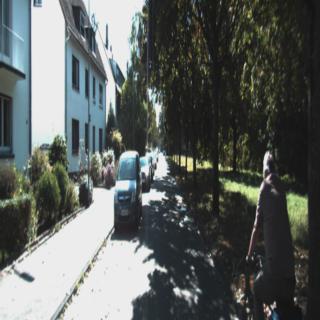}
    \includegraphics[width=0.16\columnwidth]{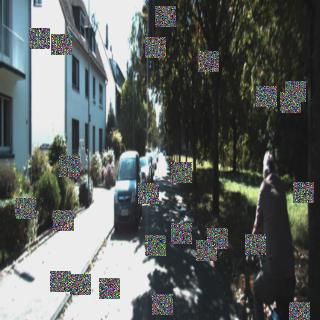}
    \includegraphics[width=0.16\columnwidth]{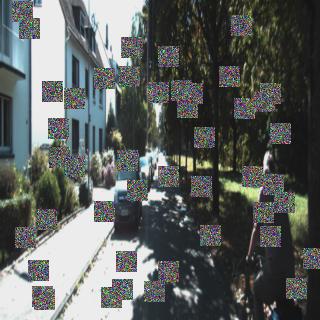}
    \includegraphics[width=0.16\columnwidth]{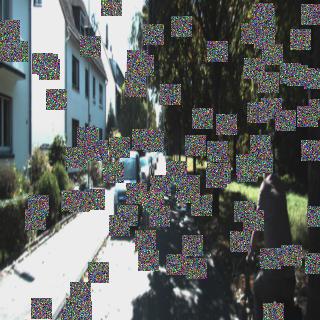}
    \includegraphics[width=0.16\columnwidth]{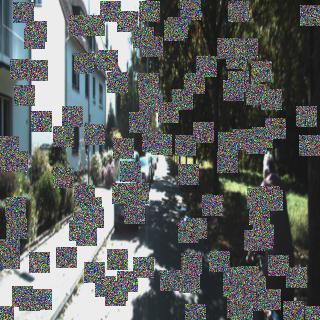}
    \includegraphics[width=0.16\columnwidth]{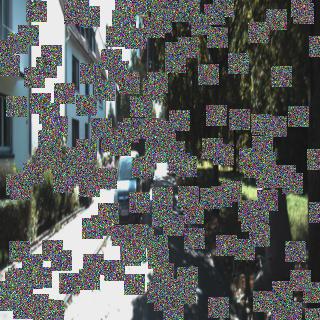}
    \caption{\textbf{Proposed augmentation method:} We show the original image and augmentation results with random 21x21 noise patches covering up to 10\%, 20\%, 40\%, 60\% and 80\% respectively.} 
    \label{fig:noise-augmentation}
\vspace{-0.2cm}
\end{figure}

Our two-stream ego-motion model is more expressive, using dense information from both appearance and depth streams, and thus tends to overfit, as we show in the ablation section of our experiments (see Section~\ref{subsec:ablation}).
However, we know that only a subset of the input pixels are reliable and needed for direct robust visual odometry~\cite{engel2018direct}, although it is challenging to estimate the optimal subset.

Consequently, we propose to leverage this insight in the form of an implicit spatial sparsity prior to regularize our ego-motion network. Our approach is based on data augmentation via sampling random noise patches to obfuscate certain parts of the input images and inferred depth maps.
This training methodology induces the following hyperparameters: (i) the percentage of the image to be covered by random noise, and (ii) the size of each noise patch. Figure~\ref{fig:noise-augmentation} shows an example of an image covered to varying degrees of obfuscation.
We present details of the effects of this data augmentation step on training and test performances in Section~\ref{subsec:ablation}, showing that it indeed reduces overfitting by learning more robust sparse features from the input images.
%
Interestingly, we also find this to be at odds with depth estimation performance, as decribed in more details in Section~\ref{subsec:ablation}.

\section{Experiments}
\label{sec:result}

\subsection{Datasets}
\label{subsec:datasets}

\begin{table}[!t]
\centering
{
\tiny
\setlength{\tabcolsep}{1.0em}
\begin{tabular}{lcccccc}
\toprule
\textbf{Method} &
Supervision & 
Snippet &
\quad Seq. 09\quad & 
\quad Seq. 10\quad \\
\midrule
SfMLearner (Zhou et al.~\cite{zhou2017unsupervised})             & Mono & 5-frame   & 0.021 $\pm$ 0.017 & 0.020 $\pm$ 0.015 \\
DF-Net~\cite{zou2018dfnet}                                       & Mono & 5-frame    & 0.017 $\pm$ 0.007 & 0.015 $\pm$ 0.009 \\
Godard et al.~\cite{godard2018digging}v3                         & Mono & 5-frame    & 0.017 $\pm$ 0.008 & 0.015 $\pm$ 0.010 \\
Klodt et al.~\cite{klodt2018supervising}                         & Mono & 5-frame    & 0.014 $\pm$ 0.007 & 0.013 $\pm$ 0.009 \\
EPC++(mono)~\cite{luo2018every}                                  & Mono & 5-frame    & 0.013 $\pm$ 0.007 & 0.012 $\pm$ 0.008 \\
GeoNet (Yin et al.~\cite{yin2018geonet})                         & Mono & 5-frame    & 0.012 $\pm$ 0.007 & 0.012 $\pm$ 0.009 \\
Struct2Depth~\cite{casser2018depth}                              & Mono & 5-frame    & 0.011 $\pm$ 0.006 & 0.011 $\pm$ 0.010 \\
Ours                                                             & Mono & 5-frame    & \textbf{0.010 $\pm$ 0.002} & \textbf{0.009 $\pm$ 0.002} \\                                                    
\midrule
Vid2Depth~\cite{mahjourian2018unsupervised}                      & Mono & 3-frame    & 0.013 $\pm$ 0.010 & 0.012 $\pm$ 0.011 \\
Shen et al~\cite{shen2019beyond}                                 & Mono & 3-frame     & \textbf{0.009 $\pm$ 0.005} & \textbf{0.008 $\pm$ 0.007} \\
\textbf{Ours}                                                             & Mono & 3-frame    & \textbf{0.009 $\pm$ 0.004} & \textbf{0.008 $\pm$ 0.007} \\                                                    
\bottomrule
\end{tabular}\\\vspace{2mm}
}
\caption{\textbf{Average Absolute Trajectory Error (ATE) in meters on the KITTI Odometry Benchmark~\cite{geiger2013vision}}: All methods are trained on Sequences 00-08 and evaluated on Sequences 09-10. The ATE numbers are averaged over all overlapping 5-frame, respectively 3-frame, snippets.}
\label{table:kitti-pose-ate}
\vspace{-0.8cm}
\end{table}


To validate our contributions we use the standard KITTI dataset~\cite{geiger2013vision}. We compare against state-of-the-art methods on the KITTI odometry benchmark which consists of 11 sequences (00-10), and we use the training protocol described by~\cite{zhou2017unsupervised}, i.e. we train on sequences 00-08 and test on sequences 09 and 10. Following related work~\cite{zhou2017unsupervised,godard2018digging}, we use the ground truth camera translation to scale our predictions; specifically, we compute the scaling factor for each two-frame prediction using a 5-frame window. We stack the two-frame predictions to obtain trajectories for each test sequence. We report the \textit{Absolute Trajectory Error (ATE)}~\cite{mur2017orb} averaged over all overlapping 3-frame and 5-frame  snippets of the test sequences as shown in Table~\ref{table:kitti-pose-ate}. In addition, we report $t_{rel}$ - average translational RMSE drift (\%) on trajectories of length 100-800m, and $r_{rel}$ - average rotational RMSE drift ($\deg/100m$) on trajectories of length 100-800m, as described by ~\cite{geiger2013vision}. We present these metrics and compare against other learning based monocular and stereo methods in Table~\ref{table:kitti-odom-traj}. In addition, in Section~\ref{subsec:ablation} we also evaluate the performance of the depth estimation component. We use the quantitative depth evaluation to illustrate the counter-intuitive dependency between jointly optimizing depth and pose. The depth evaluation is done using the Eigen test split~\cite{eigen2014depth} of the KITTI raw dataset, which consists of 697 depth images. To explore self-supervised learning of monocular SfM at scale, we consider two additional urban driving datasets in this work. We experiment with pre-training our model on the publicly available CityScapes dataset~\cite{cordts2016cityscapes} (88K images), and introduce a new urban driving dataset consisting of 24 sessions and 1 million images(the data will be made available upon request). We first filter the dataset of redundant sequential images by simply thresholding on the difference of JPEG payload size, and use the remaining images as training data. Following the pre-training step, we fine-tune on the KITTI dataset, this time using sequences 01, 02, 06, 08, 09 and 10 for training and 00, 03, 04, 05 and 07 for testing, to facilitate comparison with other methods (e.g. DVSO~\cite{yang2018deep,pillai2018superdepth}). Finally, we present our results as well as comparisons with other approaches based on direct methods, stereo, or lidar-based in Table~\ref{table:kitti-eigen-traj}.

{
\renewcommand{\textbf}{\bf}

\begin{table*}[!t]
\centering
\tiny
\setlength{\tabcolsep}{.7em}
\begin{tabular}{lccccccccccccccc}

\textbf{Method} & \textbf{Supervision} & 00$^*$  & 01$^*$  & 02$^*$  & 03$^*$  & 04$^*$  &  05$^*$ &    06$^*$  &  07$^*$  &  08$^*$   &   09$^\dagger$  & 10$^\dagger$   & \textbf{Train Avg} & \textbf{Test Avg} \\
\toprule

& \multicolumn{14}{c}{$t_{rel}$ - Average Translational RMSE drift (\%) on trajectories of length 100-800m.}  \\
\midrule
ORB-SLAM-M~\cite{mur2017orb}                  & Mono   &   25.29 &   -   &   -   &  -  &  -   &  26.01 &  -  & 24.53  &   32.40   &   -   &  - &   -   & 27.05 &   -   \\
VISO2-M~\cite{geiger2011stereoscan}              & Mono   &   18.24 &   -   &   4.37  &  -  &  -   &  19.22 &  -  & 23.61  &   24.18   &   -   &  - &   -   & 17.93 &   -   \\
SfMLearner~\cite{zhou2017unsupervised}           & Mono   & 66.4  & 35.2  & 58.8 & 10.8	 & 4.49  &  18.7 &  25.9 &  21.3 &	21.9 & 	18.8 & 	14.3 & 29.28 & 16.55 \\
Zhan et al~\cite{zhan2018unsupervised}           & Mono   &   -   &  -    &   -  &   -   &    -  &  -    &   -   &   -   &   -   &  11.9 &  12.6 &   -   & 12.30 \\
EPC++(mono)~\cite{luo2018every}                  & Mono   &   -   &  -    &   -  &   -   &    -  &  -    &   -   &   -   &   -   &  8.84 &  8.86 &   -   & 8.85  \\
UnDeepVO~\cite{li2017undeepvo}                   & Stereo & \textbf{4.14}  &	69.1  &	5.58 &	5.00 & 4.49  & 3.40  &  6.20 &  3.15 &  4.08 &  7.01 &  10.6 & 11.68 & 8.81  \\
Zhu et al~\cite{zhu2018robustness}               & Stereo & 4.95  & 45.5  & 6.40 &  4.83 & \textbf{2.43}	 & \textbf{3.97}	 &  \textbf{3.49} & \textbf{4.50}  &  \textbf{4.08} &  \textbf{4.66} &  \textbf{6.30} & 8.91 & \textbf{5.48}  \\
\textbf{Ours$\ddagger$}                                   & Mono  & 4.88   & \textbf{12.61} & \textbf{4.19} &  \textbf{4.01} &  3.2  & 5.26     &  8.18 & 6.33  &  7.34 &  6.72 &  9.52 & \textbf{6.22}  & 8.12  \\
\textbf{\textit{Ours}}                                             & \textit{Mono}   & \emph{1.29}  & \emph{1.63}  & \emph{1.06} &  \emph{1.84} & \emph{0.55}  & \emph{1.58}  &  \emph{0.91} & \emph{2.25}  &  \emph{1.84} &  \emph{3.51} &  \emph{2.32} & \emph{1.44}  & \emph{2.92} \\

\midrule
& \multicolumn{14}{c}{$r_{rel}$ - Average Rotational RMSE drift ($^{{\circ}}/100m$) on trajectories of length 100-800m.} \\ 
\midrule
ORB-SLAM-M~\cite{mur2017orb}                  & Mono   &   7.37 &   -   &   -   &  -  &  -   &  10.62 &  -  & 10.83  &   12.13   &   -   &  - &   -   & 10.23 &   -   \\
VISO2-M~\cite{geiger2011stereoscan}              & Mono   &    2.69 &   -   &   1.18  &  -  &  -   &  3.54 &  -  & 4.11  &   2.47   &   -   &  - &   -   & 2.80 &   -   \\
SfMLearner~\cite{zhou2017unsupervised}           & Mono   & 6.13& 2.74  & 3.58 & 3.92 & 5.24 & 4.1  & 4.8  & 6.65 & 2.91 & 3.21 & 3.30 & 4.45 & 3.26\\
Zhan et al~\cite{zhan2018unsupervised}           & Mono   &   - &  -    &   -  &   -  &   -  &  -   &   -  &   -  &   -  & 3.60 & 3.43 &   -  & 3.52 \\         
EPC++(mono)~\cite{luo2018every}                  & Mono   &   - &  -    &   -  &   -  &   -  &  -   &   -  &   -  &   -  & 3.34 & 3.18 &   -  & 3.26 \\         
UnDeepVO~\cite{li2017undeepvo}                   & Stereo & 1.92& 1.60	& 2.44 & 6.17 & 2.13 & 1.5	& 1.98 & 2.48 &	1.79 & 3.61 & 4.65 & 2.45 & 4.13 \\
Zhu et al~\cite{zhu2018robustness}               & Stereo   & 1.39& 1.78	& 1.92 & 2.11 &	1.16 & 1.2  & 1.02 & 1.78 & 1.17 & 1.69 & 1.59 & 1.50 & 1.64 \\
\textbf{Ours}                                             & Mono   & \bf{0.55}& \bf{0.48}  & \bf{0.45} & \bf{0.94} & \bf{0.45} & \bf{0.67} & \bf{0.34} & \bf{1.15} & \bf{0.70} & \bf{1.57} & \bf{1.48} & \bf{0.64} & \bf{1.53}\\

\bottomrule
\end{tabular} \\

\caption{Comparison to self-supervised learning methods on the KITTI odometry benchmark. We report the following metrics: $t_{rel}$ - average translational RMSE drift (\%) and $r_{rel}$ - average rotational RMSE drift ($^{{\circ}}/100m$). The methods are trained on Sequences 00-08 ($^*$) and tested on Sequences 09 and 10 ($^\dagger$); The results of the methods trained with monocular data were scaled using the scale from the ground truth translation. $\ddagger$ denotes global scale alignment while \textit{italics} denote iterative scaling of snippet trajectories. The numbers for~\cite{geiger2011stereoscan,mur2017orb,li2017undeepvo} are reported from~\cite{li2017undeepvo}.}
\label{table:kitti-odom-traj}
\vspace{0mm}
\end{table*}
}

\subsection{Implementation details}
\label{subsec:implementation}

All our models are implemented in PyTorch~\cite{paszke2017automatic}, using the Adam optimizer~\cite{kingma2014adam}  with $\beta_1$ = 0.9 and $\beta_2$ = 0.999. Both the depth and the pose networks receive as input images of size $320 \times 320$ pixels. We set the SSIM weight $\alpha=0.85$ and the depth smoothness weight $\lambda=0.1$. We use a batch size of 8 images and train for 200 epochs. We start with learning rates of $1e^{-3}$ and $5e^{-4}$ for the depth and pose networks respectively, and reduce the learning rates by a factor of $0.5$ every 80 epochs. For the regularization and sparsity-inducing data augmentation, we performed an ablation study (more details in Section~\ref{subsec:ablation}) and found a threshold of $20\%$-$40\%$ of the image size and square blocks with sides of 80-100 pixels to work best. 

\subsection{Results on the KITTI odometry benchmark}
\label{subsec:KITTI_odometry}

Table~\ref{table:kitti-pose-ate} shows our ATE results on Sequences 09 and 10, evaluated on 3 and 5-frame snippets. As can be seen, our method achieves state-of-the-art results compared to other learning-based methods, and on par with~\citet{shen2019beyond}. While~\citet{shen2019beyond} propose to augment the photometric error between two frames through an additional loss term imposed via geometric constraints derived from epipolar geometry, our work shows that the early fusion of depth and RGB information coupled with an appropriate training scheme can achieve competitive results without the need of additional loss terms or external information. In addition, we also report the average translational RMSE drift ($t_{rel}$) and average rotational RMSE drift ($r_{rel}$) on trajectories of length 100-800m in Table~\ref{table:kitti-odom-traj}. We stack our unscaled predictions to form the complete trajectory, and compute a global alignment factor~\cite{grupp2017evo}; these results are denoted by $\ddagger$ in Table~\ref{table:kitti-odom-traj}. In addition, in \textit{italics} we report results when the scale factor is computed as described in Section~\ref{subsec:datasets}, i.e. we compute the scale for our incremental predictions over 5-frame snippets and stack the scaled incremental predictions to form the complete trajectory. 

Our $t_{rel}$ and $r_{rel}$ results are better than related monocular or stereo based learning methods, even though we estimate the transformations on a frame-to-frame basis and do not explicitly account for the errors induced by dynamic objects or occlusions when computing the photometric loss. Our model achieves state-of-the-art results when compared to similar methods due to the use of two complementary modalities: appearance and geometry. In addition, by explicitly regularizing our model during training using the proposed method, we also force the model to learn sparse features which further increase its performance. We note that our globally aligned trajectories fall short of~\citet{zhu2018robustness} in terms of the $t_{rel}$ metric - our method suffers from the standard scale drift present in monocular self-supervised methods, while~\cite{zhu2018robustness} use stereo supervision for training as well as RANSAC to filter outliers and the pose estimate is computed externally and not learned. Interestingly, the $t_{rel}$ gap between the iteratively scaled and globally scaled results is narrowed significantly when more data is used for training, as described in Table~\ref{table:kitti-eigen-traj}. We present qualitative results of our method on the test Sequences 09 and 10 in Figure~\ref{fig:supplementary-pose-trajectory-test}, with more qualitative results on the training sequences in the supplementary materials. 

\begin{figure}[t!]
    \centering
    \includegraphics[height=4.5cm]{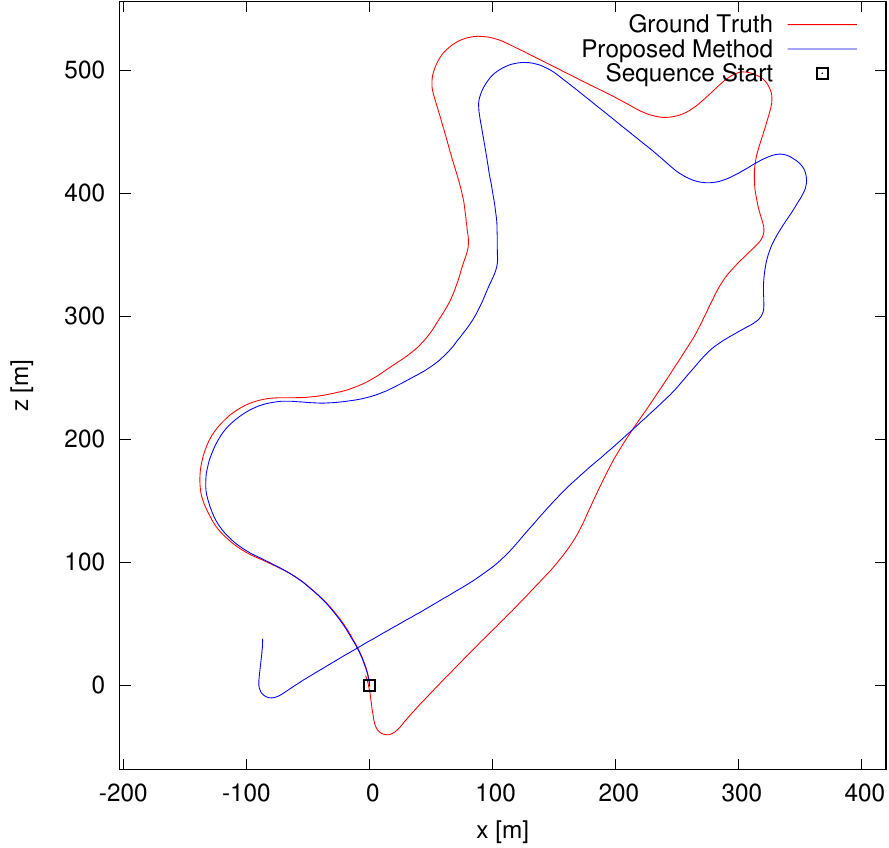}\quad\quad
    \includegraphics[height=4.5cm]{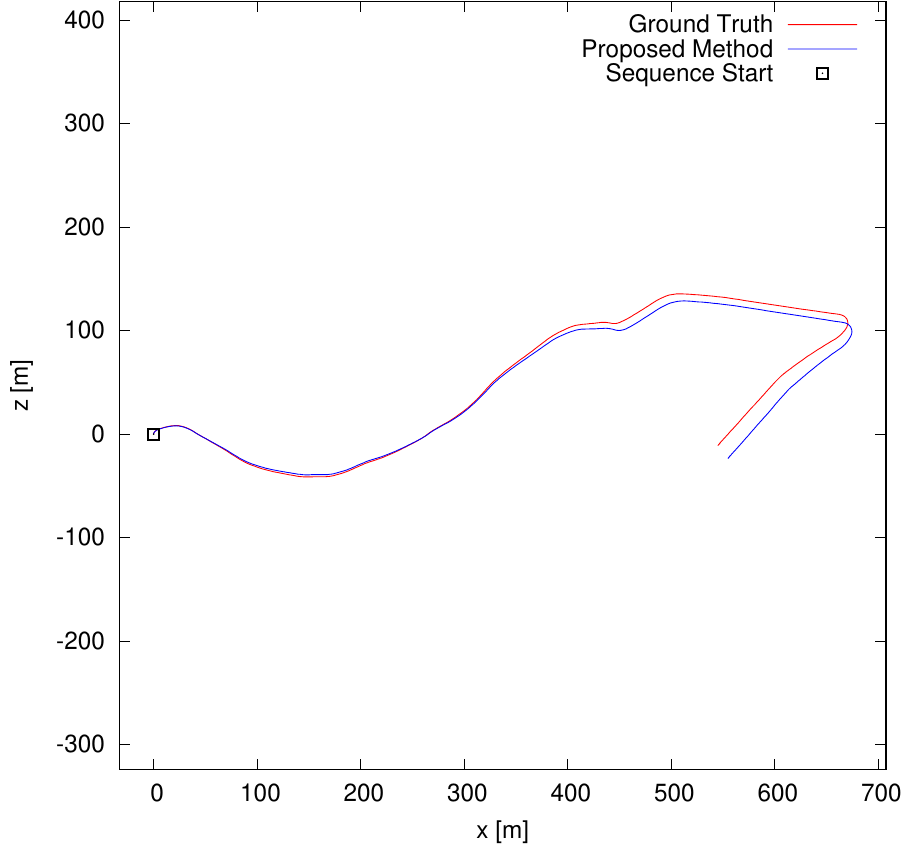}
    \caption{Qualitative trajectory results of the proposed method on test sequences 09 and 10 of the KITTI odometry benchmark.}
    \label{fig:supplementary-pose-trajectory-test}
    \vspace*{-5mm}
\end{figure}

\subsection{Ablation study}
\label{subsec:ablation}

\begin{table}[!b]
\vspace{-0.4cm}
\centering
\tiny
\setlength{\tabcolsep}{.7em}
\begin{tabular}{l|cc|cc|ccccc}


& \multicolumn{4}{c}{\textbf{Pose}} & \multicolumn{5}{c}{\textbf{Depth}} \\
 & \multicolumn{2}{c}{Train} & \multicolumn{2}{c}{Test} & Abs Rel & Sq Rel & RMSE & RMSE log & $\delta < 1.25$ \\ 
\toprule
Method & $t_{rel}$ & $r_{rel}$ & $t_{rel}$ & $r_{rel}$ \\
{Ours} - RGB, w/o Depth, w/o aug      & 1.39 & 0.59 & 5.93 & 2.64 & 0.138 &1.084&5.336 & 0.220&0.823 \\
{Ours} - RGB + aug, w/o Depth         & 2.61 & 1.10 & 4.62 & 2.15 & 0.143 &1.110&5.335 & 0.220&0.811 \\
{Ours} - RGB + Depth, w/o aug         & 1.16 & 0.43 & 4.91 & 2.44 & 0.135 &1.031&5.260 & 0.216&0.826 \\

{Ours}                                & 1.44 & 0.64 & 2.92 & 1.53 & 0.139&1.063 &5.349 &0.221 &0.817 \\   

\bottomrule
\end{tabular} \\

\caption{\textbf{Ablation study on the KITTI odometry benchmark.} We report $t_{rel}$ \& $r_{rel}$ for different versions of our method. All methods are trained on Sequences 00-08, tested on Sequences 09 \& 10.}
\label{table:kitti-odom-ablation}
\end{table}

We present ablation results in Table~\ref{table:kitti-odom-ablation}. The first row represents our baseline, which uses only RGB images as input.
The third row shows results using the proposed two-stream network architecture, described in Section~\ref{subsec:pose_network}, but without the proposed regularization and sparsity-inducing training methodology. As we can see, the increased modeling power of the pose network leads to even more overfitting during training and increased depth performance at test time. This shows that when optimizing for depth we are effectively interested in letting the pose network overfit as much as possible, at the expense of its test-time generalization performance. Rows 2 and 4 of Table~\ref{table:kitti-odom-ablation} show the performance of the baseline and of the proposed two-stream network architecture with the proposed training methodology: the ego-motion network overfits less at train time, which leads to much better performance at test time. Evaluating depth shows that due to reduced performance of the ego-motion component during training, the performance of the depth network suffers as well. 

\begin{figure}[t!]
    \centering
    {
    \subfloat{
    \includegraphics[ width=0.4\linewidth]{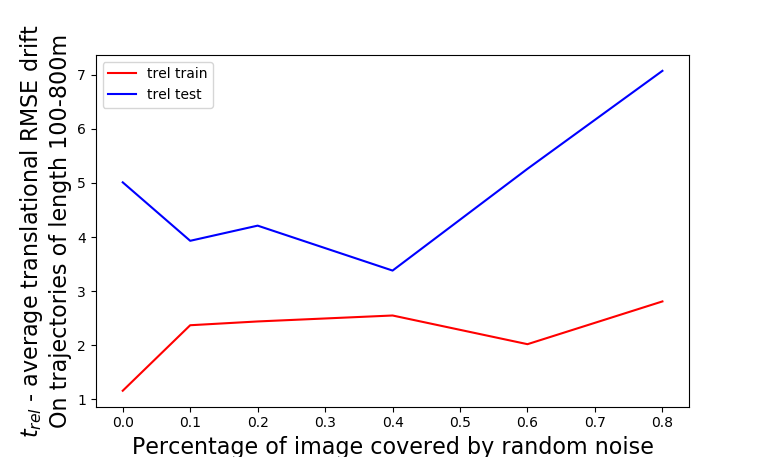}
    \label{fig:trel_avg}}\quad
    \subfloat{
    \includegraphics[ width=0.4\textwidth]{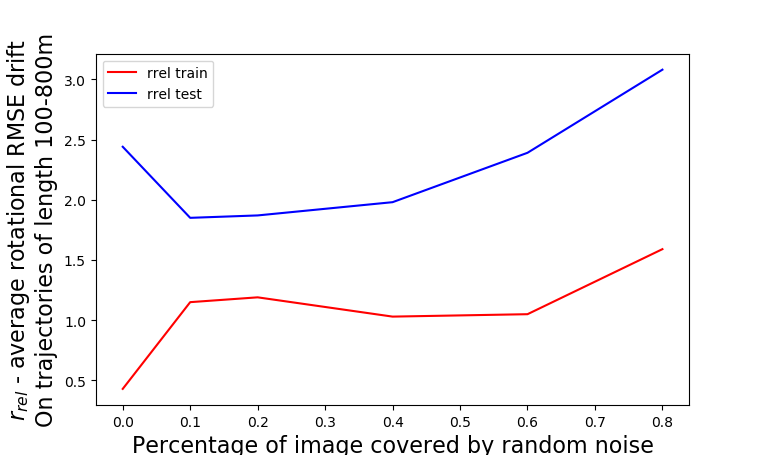}
    \label{fig:rrel_avg}}
    \quad
    }
    \caption{\textbf{Ablation study on the amount of noise augmentation:} We show test and train $t_{rel}$ (left) and $r_{rel}$ (right) computed when training with varying levels of the proposed augmentation method. We obtain the best results when covering $20\%$ of the input images with random noise. }
    \label{fig:noise_aumgentation_ablation}
\vspace{-0.3cm}
\end{figure}

We also perform an ablative analysis to better understand how our proposed augmentation technique affects performance.
Specifically, we vary 2 hyperparameters: the percentage of pixels to cover with noise ($10\%$, $20\%$, $40\%$, $60\%$, $80\%$) and the size noise patches to apply (we experiment with square patches of size 21x21, 41x41, 61x61, 81x81 and 101x101 pixels, see Figure~\ref{fig:noise-augmentation}). The results are summarized in Figure~\ref{fig:noise_aumgentation_ablation}: the proposed augmentation technique has the desired effect of regularizing the solution and reducing overfitting. We get the best test results when employing an augmentation-level of $20\%$-$40\%$, and when using the larger size patches (81x81 and 101x101). When going above $40\%$ augmentation the solution degrades; however we note that, surprisingly, even when replacing $80\%$ of the input image with random noise we are still able to regress the ego-motion.

\subsection{Does self-supervised learning of ego-motion improve with more data?}
\label{subsec:large_scale}

To further evaluate our method, we study the effects of self-supervised pre-training on large datasets. We introduce an urban driving dataset of 1 million frames, and show that by pre-training the network with additional data we are able to gradually improve monocular pose estimation performance on a target dataset such as KITTI~\cite{geiger2013vision}. Furthermore, we compare these results to pre-training on CityScapes~\cite{cordts2016cityscapes}, a dataset similar to KITTI. We show that large amounts of unlabeled driving data can be a scalable alternative to highly curated datasets in the same domain.


As evidenced by previous works~\cite{godard2017unsupervised}, pretraining on the CityScapes dataset~\cite{cordts2016cityscapes} (CS) can substantially improve depth and pose estimation performance on the KITTI dataset. In addition, in this work we also investigate how these models scale and perform when self-supervised with larger amounts of unlabelled video data. For all self-supervised pre-training experiments, we first pretrain our models either on the approximately 80K images of CityScapes (CS) or the 1M urban driving dataset (D1M) (i.e. approximately 50 epochs on CS or 5 epochs on the D1M driving dataset). Subsequently, we fine-tune the model on the target KITTI dataset, using the same protocol as described in Section~\ref{subsec:implementation}. For a fair comparison we use the KITTI odometry sequences \textit{01, 02, 06, 08, 09, 10} for training and test on sequences \textit{00, 03, 04, 05, 07}. When training with the 1M dataset, we notice the benefits of self-supervision at scale and observe noticeable improvements in pose estimation performance. Interestingly,  CityScapes (CS) pre-training performs quite well in the ego-motion benchmark despite having pre-trained on a smaller dataset compared to the 1M driving dataset (D1M). This can be attributed to the similarity in domains between CityScapes (CS) and the target KITTI dataset that were both captured in geographically similar regions. These results, as well as comparisons against direct, traditional and lidar based methods are summarized in Table~\ref{table:kitti-eigen-traj}. As in Table~\ref{table:kitti-odom-traj}, we denote our results when computing a global alignment step with $\ddagger$. As before, we notice a performance drop in the $t_{rel}$ metric which we ascribe to scale inconsistency of our self-supervised model. However, we record a significant improvement in the $t_{rel}$ metric with data for the globally scaled models, from which we conclude that training on more data has the additional effect of regularizing the scale of the model across the dataset. We note that our method falls short of the state-of-the-art results obtained by direct methods such as DVSO~\cite{yang2018deep} or ORB-SLAM2~\cite{mur2017orb} (i.e. without loop closure and global optimization), however, we outperform the learning-based methods of~\citet{pillai2018superdepth,wang2017deepvo} and our orientation estimates also outperform~\citet{geiger2011stereoscan}.

\begin{table*}[!t]
\centering
\tiny
\setlength{\tabcolsep}{.7em}
\begin{tabular}{lcccccccccccccc}

\textbf{Method} & \textbf{Sensor} & 01$^*$ &    02$^*$  &  06$^*$  &  08$^*$   &   09$^*$  & 10$^*$   & 00$^\dagger$  & 03$^\dagger$  & 04$^\dagger$  & 05$^\dagger$  & 07$^\dagger$  & \textbf{Train Avg} & \textbf{Test Avg} \\
\toprule

& \multicolumn{12}{c}{$t_{rel}$ - Average Translational RMSE drift (\%) on trajectories of length 100-800m.}  \\
\midrule
SuperDepth~\cite{pillai2018superdepth}           & Stereo & 13.48 & 3.48 & 1.81  & 2.25  &  3.74 &  2.26 & 6.12  & 7.90  & 11.80 &  4.58 & 7.60  & 4.50  & 7.60  \\
DeepVO~\cite{wang2017deepvo}                    &Mono+Pose&   -   &   -   & 5.42 &   -   &   -   &8.11   &   -   & 8.49 &  7.19 &  2.62 &  3.91  &   -   & 5.96  \\
Velas et al.~\cite{velas2018cnn}                &Lidar+IMU& 4.44  & 3.42 & 1.88  & 2.89  &  4.94 & 3.27  & 3.02  & 4.94  & 1.77  &  2.35 & 1.77  &  3.11 & 3.22  \\ 
VISO2\_S~\cite{geiger2011stereoscan}             & Stereo &   -   &   -   & 1.48 &   -   &   -   &  1.17 &  -    & 3.21  & 2.12  &  1.53 &  1.85 &  -    & 1.89  \\
LO-NET~\cite{li2019net}                          & Lidar  &  1.36 & 1.52  & \textbf{0.71} &  2.12 &  1.37 &  1.80 & 1.47  & 1.03  &  0.51 &  1.04 &  1.70 & 1.09  & 1.75  \\
LOAM~\cite{zhang2017low}                         & Lidar  &  \textbf{0.78} &  1.43 & 0.92 &  \textbf{0.86} &  \textbf{0.71} &  \textbf{0.57} & \textbf{0.65}  &  \textbf{0.63} &  1.12 &  0.77 &  0.79 &  -    &  1.33  \\ 
ORB-SLAM2~\cite{mur2017orb}                      & Stereo  &  1.38 &  \textbf{0.81} & 0.82 &  1.07 &  0.82 &  0.58 & 0.83  &  0.71 &  0.45 &  0.64 &  0.78 &  -    & 0.81  \\
DVSO~\cite{yang2018deep}                         & Stereo &  1.18 &  0.84 & \textbf{0.71} &  1.03 &  0.83 &  0.74 & 0.71  &  0.77 &  \textbf{0.35} &  \textbf{0.58} &  \textbf{0.73} & \textbf{0.89}  & \textbf{0.63}  \\
\midrule
\textbf{Ours - KITTI$\ddagger$}                           & Mono   & 17.59 &  6.82 & 8.93 &  8.38 &  6.49 & 9.83 &  7.16 &  7.66 & 3.8   &   6.6 &  11.48 & 9.67 &  7.34 \\
\textbf{Ours - KITTI + CS$\ddagger$}                      & Mono   & 11.36 & 3.41  & 6.41 & 6.67  &  4.48 & 8.84 &  5.85 &  5.99 & 2.69 &  4.79 & 7.06 & 6.86 & 5.28 \\
\textbf{Ours - KITTI + D1M$\ddagger$}                     & Mono   & 9.04  &  5.15 & 4.21 &  5.07 &  3.7  & 6.9  &  5.42 &  6.92 &  2.87  & 5.07  & 4.28 & 5.68  &  4.91 \\
\textbf{\textit{Ours - KITTI}}                                     & \textit{Mono}   &  \textit{4.74} &  \textit{2.6}  & \textit{0.97} &  \textit{1.72} &  \textit{1.98} &  \textit{2.56} & \textit{3.83} &   \textit{5.74} &  \textit{1.45} &  \textit{1.54} &  \textit{2.94} &  \textit{2.43} &  \textit{3.1}  \\
\textbf{\textit{Ours - KITTI + CS}}                          & \textit{Mono}   &  \textit{1.45} & \textit{1.32}  & \textit{0.78} &  \textit{1.87} &  \textit{1.20} &  \textit{1.14} & \textit{3.55} &   \textit{4.19} &  \textit{1.52} &  \textit{2.29} &  \textit{3.08} &  \textit{1.29} &  \textit{2.93} \\
\textbf{\textit{Ours - KITTI + D1M}}                              & \textit{Mono}   &  \textit{0.91} &  \textit{1.22} & \textit{0.79} &  \textit{1.57} &  \textit{1.28} &  \textit{0.84} & \textit{3.04} &   \textit{3.88} &  \textit{1.40} &  \textit{2.16} &  \textit{2.57} &  \textit{1.10} &  \textit{2.61} \\
\midrule
& \multicolumn{12}{c}{$r_{rel}$ - Average Rotational RMSE drift ($^{{\circ}}/100m$) on trajectories of length 100-800m.} \\ 
\midrule
SuperDepth~\cite{pillai2018superdepth}           & Stereo & 1.97 & 1.10 & 0.78 & 0.84 & 1.19 & 1.03 & 2.72& 4.30  & 1.90 & 1.67 & 5.17 & 1.15 & 3.15 \\
DeepVO~\cite{wang2017deepvo}                    &Mono+Pose&   -  &  -   & 5.82 &   -  &   -  & 8.83 &  -   & 6.89 & 6.97 & 3.61 & 4.60 &  -  & 6.12 \\
VISO2\_S~\cite{geiger2011stereoscan}             & Stereo &   -  &   -  & 1.58 & -    &   -  & 1.30 & -    & 3.25 & 2.12 & 1.60 & 1.91 &  -   &  1.96\\
LO-NET~\cite{li2019net}                          & Lidar  & 0.47 & 0.71 & 0.50 & 0.77 & 0.58 & 0.93 & 0.72 & 0.66 & 0.65 & 0.69 & 0.89 &  0.63& 0.79 \\
ORB-SLAM2~\cite{mur2017orb}                      & Stereo & 0.20 & 0.28 & 0.25 & 0.31 & 0.25 & 0.28 & 0.29 & 0.17 & 0.18 & 0.26 & 0.42 & -   & 0.26 \\
DVSO~\cite{yang2018deep}                         & Stereo & \textbf{0.11} & \textbf{0.22} & \textbf{0.20} &
\textbf{0.25} & \textbf{0.21} & \textbf{0.21} & \textbf{0.24} & \textbf{0.18} & \textbf{0.06} & \textbf{0.22} & \textbf{0.35} & \textbf{0.20} & \textbf{0.21} \\
\midrule
\textbf{Ours - KITTI}                                     & Mono   & 1.01 & 0.87 & 0.39 & 0.61 & 0.86 & 0.98 & 1.70 & 3.49 & 0.42 & 0.90 & 2.05 & 0.79 & 1.71\\
\textbf{Ours - KITTI + CS}                                & Mono   & 0.59 & 0.57 & 0.46 & 0.66 & 0.47 & 0.56 & 1.55 & 2.82 & 0.78 & 1.08 & 1.97 & 0.55 & 1.64 \\
\textbf{Ours - KITTI + D1M}                               & Mono   & 0.35 & 0.56 & 0.39 & 0.58 & 0.48 & 0.58 & 1.32 & 2.97 & 0.62 & 1.04 & 1.75 & 0.49 & 1.54 \\
\bottomrule
\end{tabular}

\caption{Comparison to direct, feature based, lidar based and other learned methods on the KITTI odometry benchmark. We report the following metrics $t_{rel}$ and $r_{rel}$ averaged over trajectories of length 100-800m. $\dagger$ and $*$ represent test and respectively train seq. for our method, as well as for~\cite{pillai2018superdepth, yang2018deep}. \cite{li2019net} and~\cite{velas2018cnn} are trained on Seq. 00-06 and tested on Seq. 07-10. DeepVO~\cite{wang2017deepvo} is trained on Seq. 00, 02, 08 and 09. The numbers for~\cite{velas2018cnn} are taken from~\cite{li2019net}. The results of the methods trained only with Mono data were scaled using ground truth depth scale. $\ddagger$ denotes global scale alignment while \textit{italics} denote iterative scaling of snippet trajectories. Our self-supervised method is not yet competitive with stereo and lidar, but shows a clear trend of improvement with more data, towards closing the gap with these complex methods.}
\label{table:kitti-eigen-traj} 
\vspace{-5mm}
\end{table*}


\section{Conclusion}
\label{sec:conclusion}

This paper addresses the problem of learning monocular ego-motion estimation in a self-supervised setting. We explore the inter-dependence between depth regression and ego-motion estimation in the self-supervised regime, gaining insights into training methodologies when optimizing for ego-motion. Leveraging our insights, we propose a new two-stream network architecture along with a sparsity-inducing image augmentation technique that reduces pose overfitting, allowing the network to better generalize. We validate our contributions through extensive comparisons on the standard KITTI benchmark and we show that our method achieves state-of-the-art results. In addition, we also investigate the ability of self-supervised learning methods to scale with unlabeled data by training our method on an urban driving dataset containing 1 million images. We show that through self-supervised pre-training we are able to achieve additional gains, further narrowing the performance gap between learned and direct ego-motion estimation methods. 
\clearpage
\appendix

\section{Depth estimation network architecture}

\begin{figure}[b]
	\centering
	{
		\includegraphics[ width=1.0\linewidth, height=0.44\linewidth]{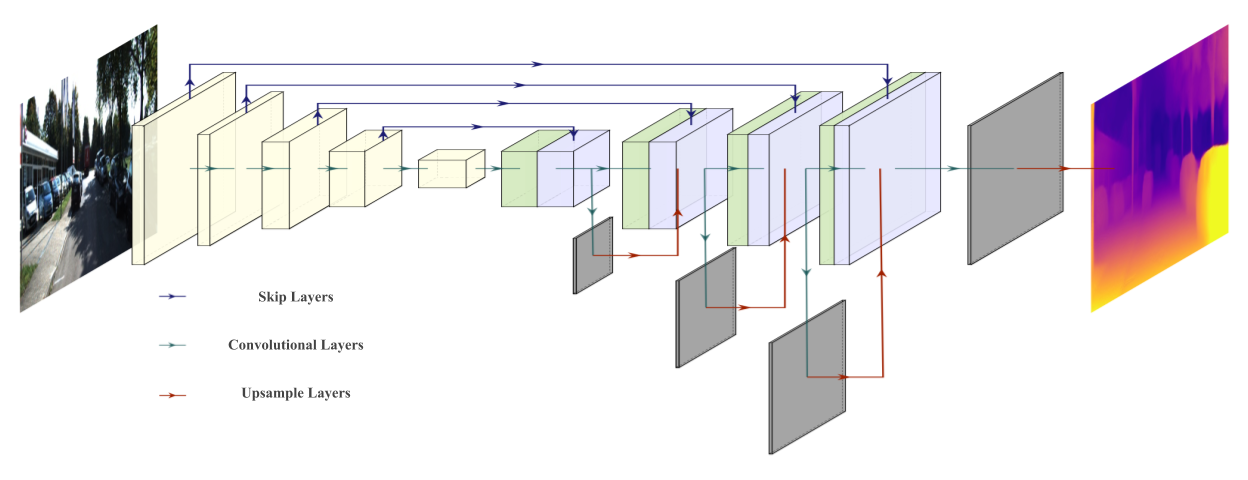}
	}
	\caption{The architecture of our depth estimation network.}
	\label{fig:depth-architecture}
\end{figure}

We show in Fig.~\ref{fig:depth-architecture} the architecture of the depth network used. We base our architecture on~\cite{mayer2016large} and follow~\cite{pillai2018superdepth} to add skip connections and output depth at 4 scales. 

\begin{figure}[b!]
    \centering
    \includegraphics[height=4cm]{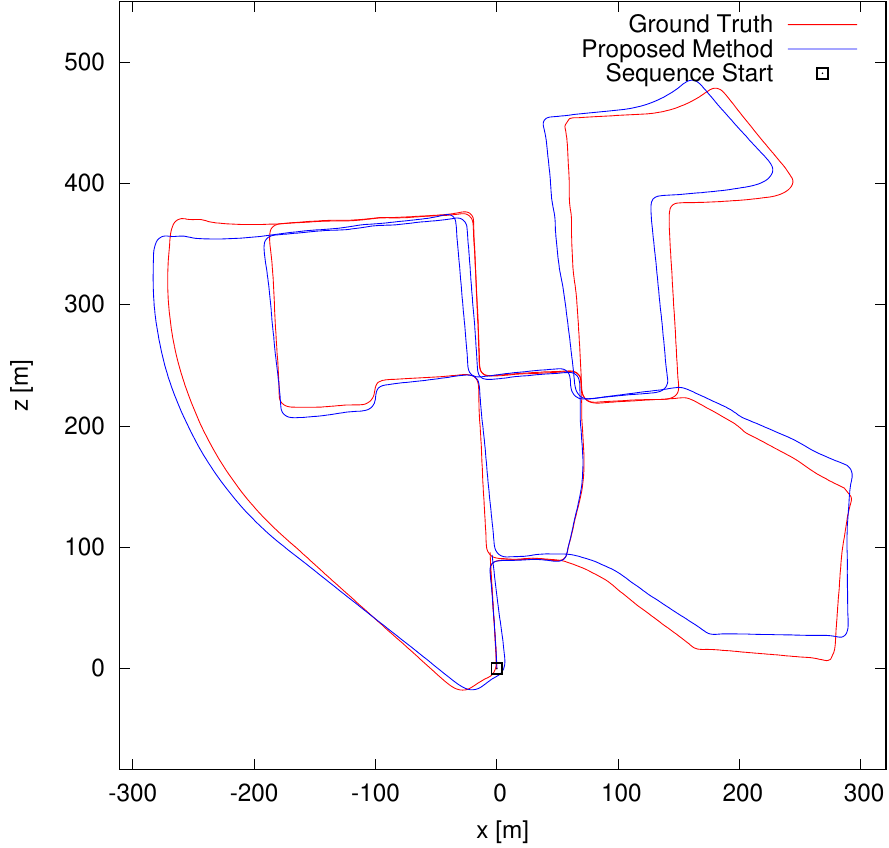}
    \includegraphics[height=4cm]{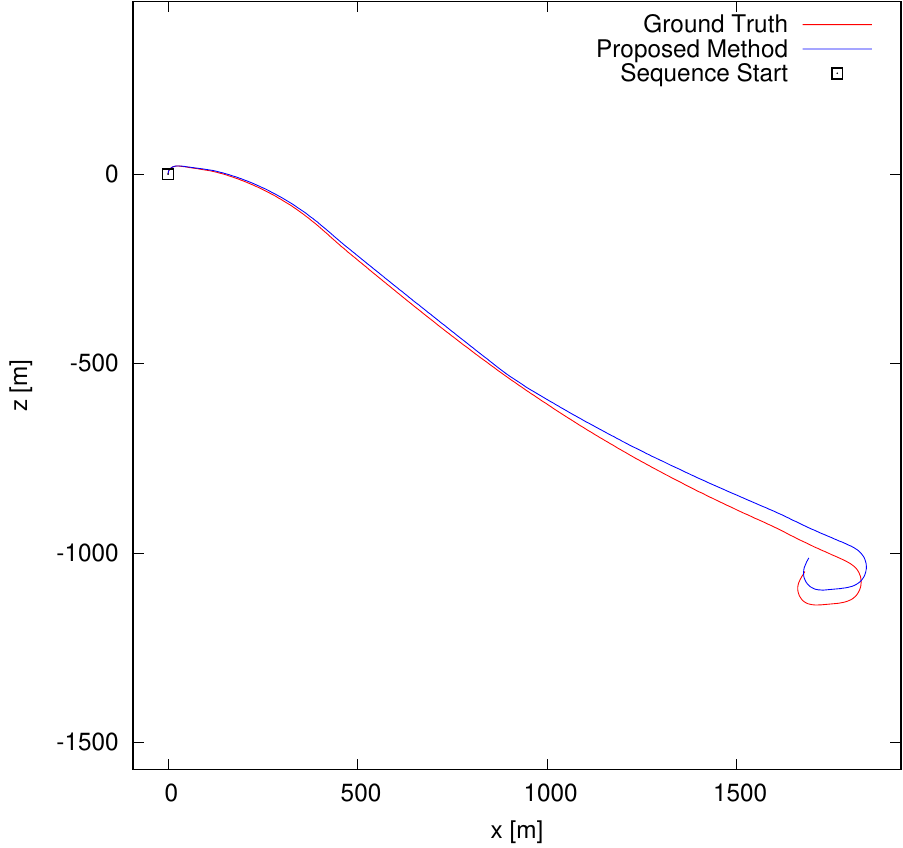}
    \includegraphics[height=4cm]{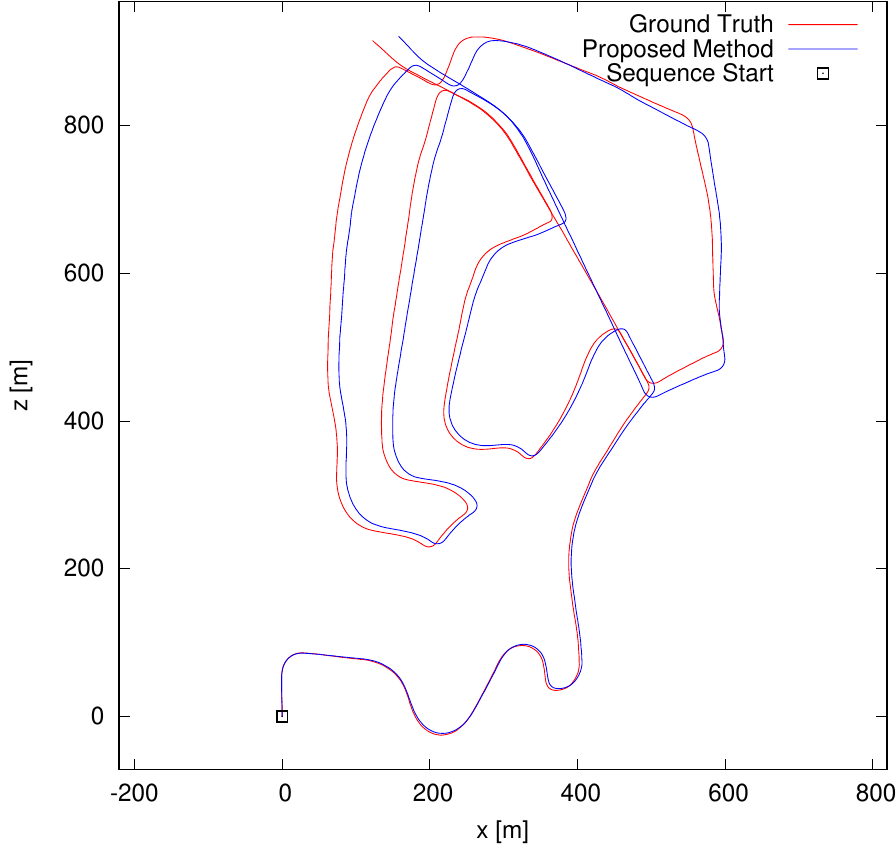} \\
    \includegraphics[height=4cm]{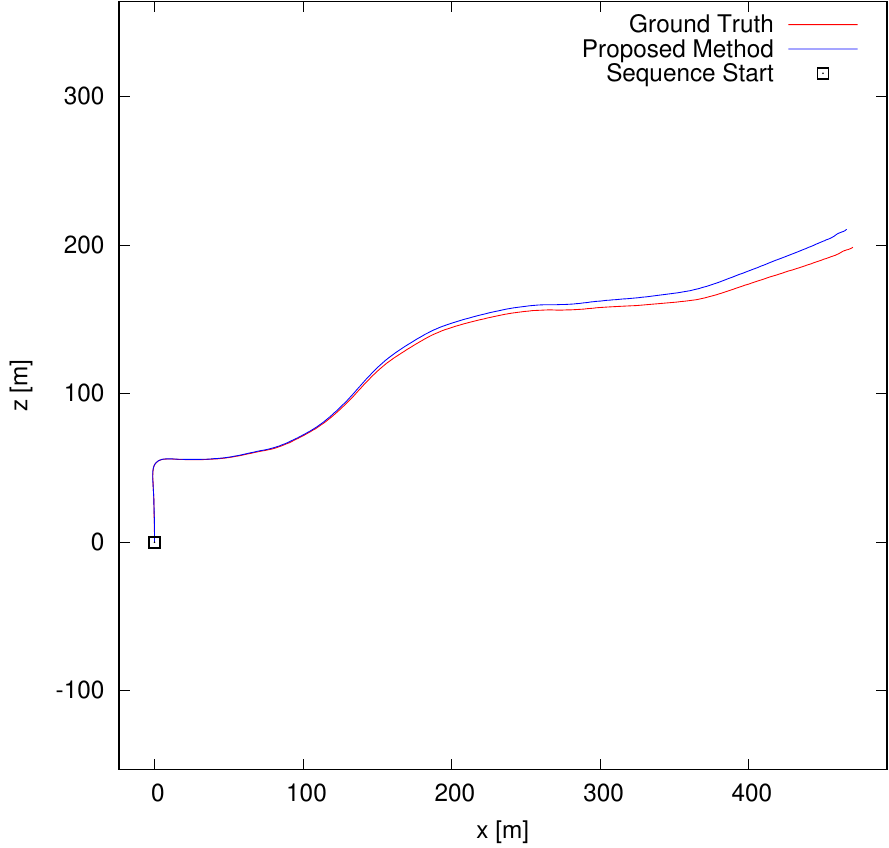}
    \includegraphics[height=4cm]{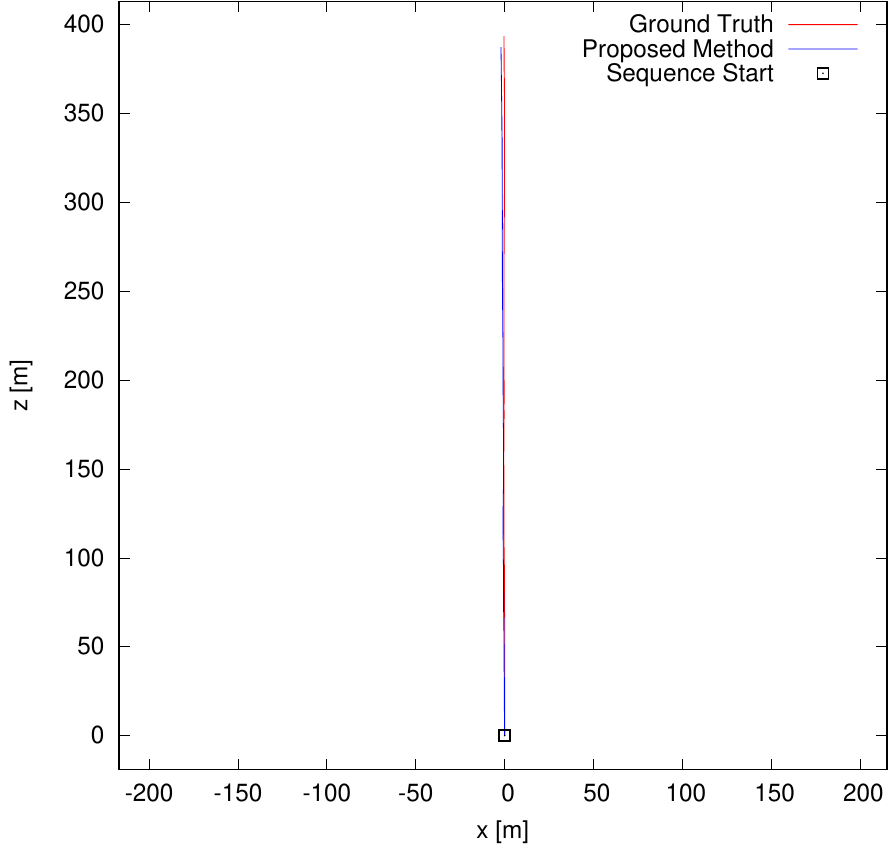}
    \includegraphics[height=4cm]{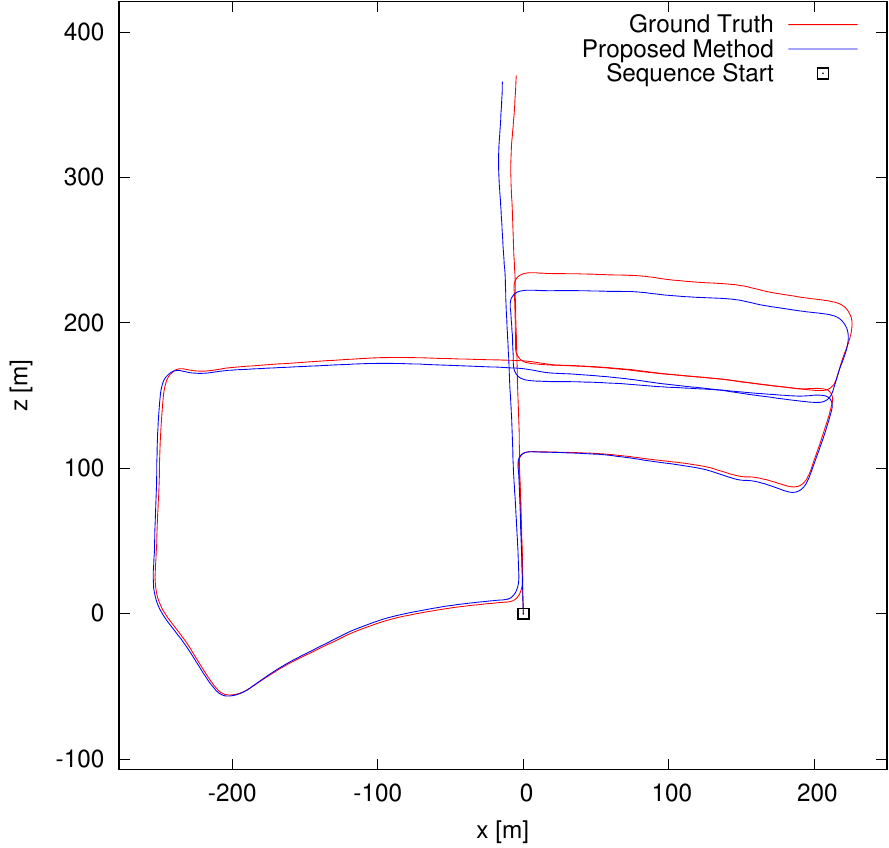} \\
    \includegraphics[height=4cm]{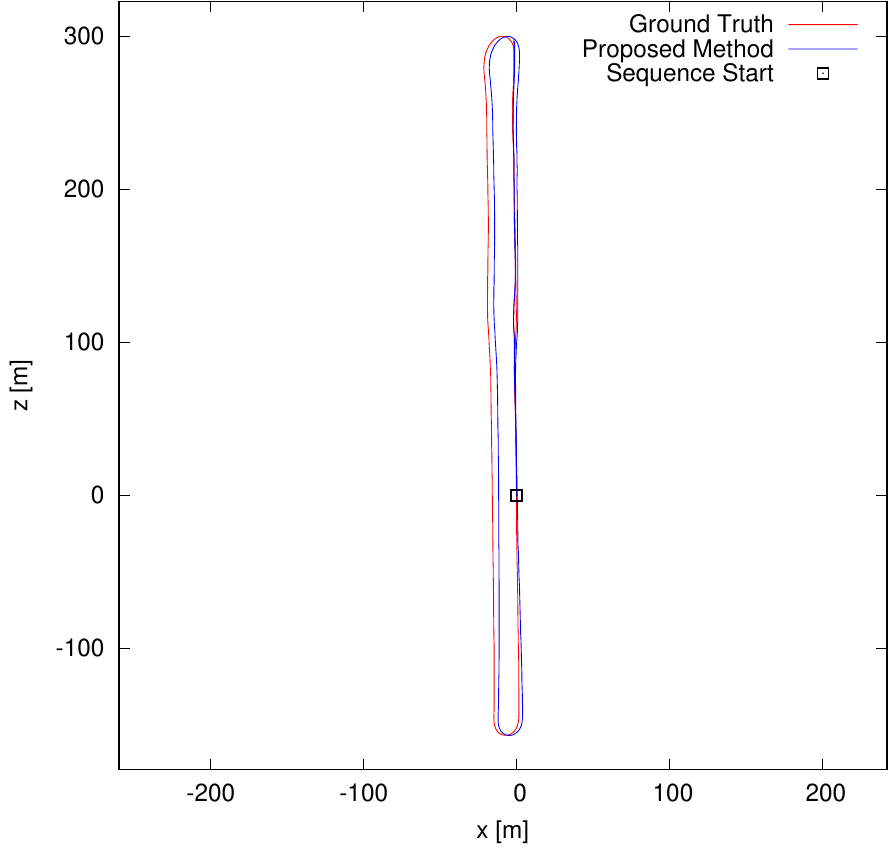}
    \includegraphics[height=4cm]{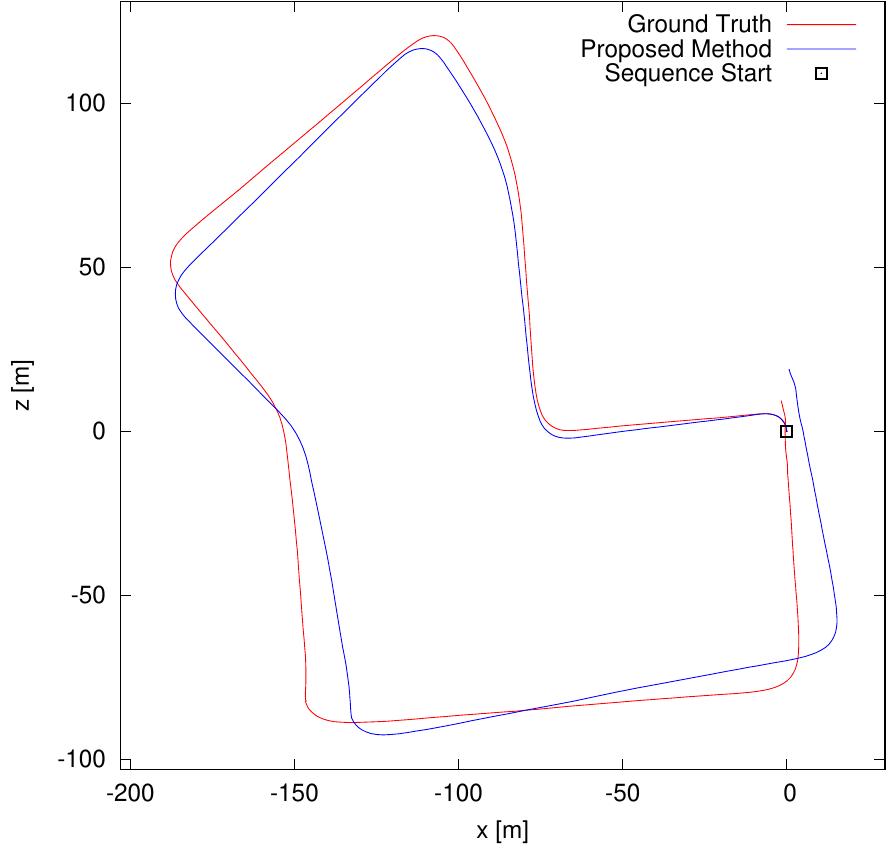}
    \includegraphics[height=4cm]{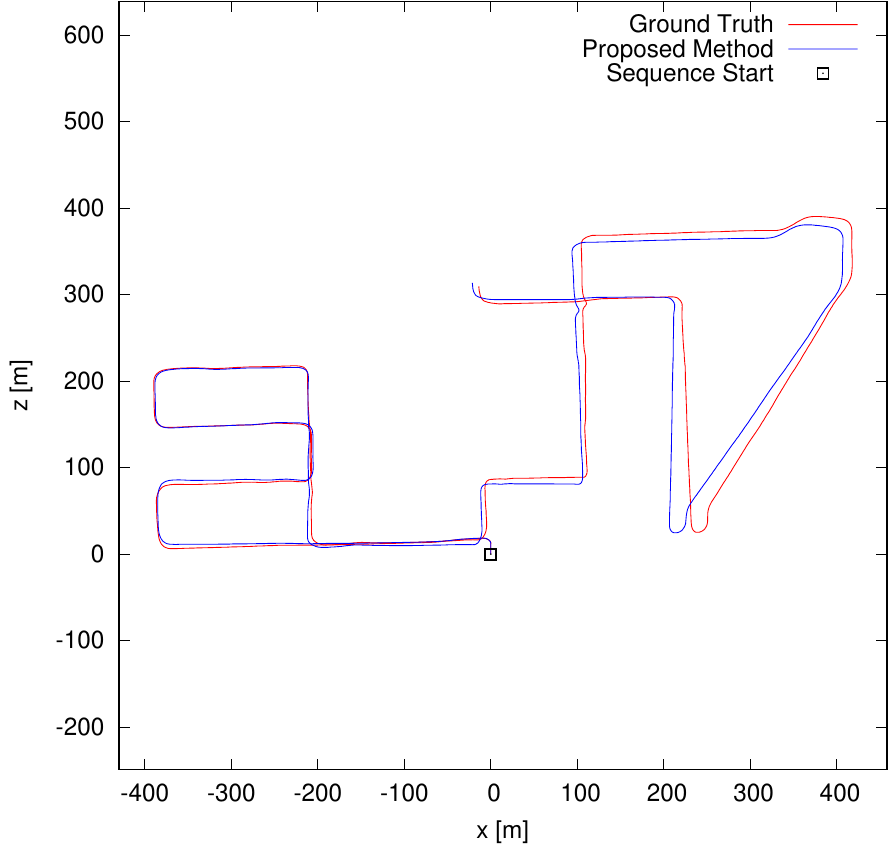}
    \caption{Qualitative trajectory results of the proposed method on train sequences 00-08 of the KITTI odometry benchmark.}
    \label{fig:supplementary-pose-trajectory-train}
\end{figure}

\section{Qualitative Results}
\label{sec:supplementary-qualitative-results}


We present qualitative results of our method on the training sequences 00-08 of the KITTI~\cite{geiger2013vision} odometry benchmark in Figure~\ref{fig:supplementary-pose-trajectory-train}.


\section{Structural Similarity (SSIM) loss component}

As described in~\cite{wang2004image}, the SSIM loss between two images is defined as:

\begin{equation}
    SSIM(\mathbf{x},\mathbf{y}) = \frac{(2\mu_{x}\mu_{y} + C_1)(2\sigma_{xy} + C_2)}{(\mu_{x}^{2} + \mu_{y}^{2} + C_1)(\sigma_{x}^{2} + \sigma_{y}^{2}+C_2)}
  \label{eq:loss-robust}
\end{equation}

In all our experiments $C_1=1e^{-4}$ and $C_2=9e^{-4}$, and we use a $3x3$ block filter to compute $\mu_x$ and $\sigma_x$ - the per-patch mean and standard deviation.

\small
\bibliography{references}  


\end{document}